\pdfoutput=1

\documentclass[11pt]{article}

\usepackage[]{acl}

\usepackage{times}
\usepackage{latexsym}

\usepackage[T1]{fontenc}

\usepackage[utf8]{inputenc}

\usepackage{microtype}

\usepackage{times}
\usepackage{latexsym}

\usepackage{microtype,comment}
\usepackage{graphicx}
\usepackage{caption,subcaption}
\usepackage{multirow,epstopdf}
\usepackage{makecell}
\usepackage{amsmath,natbib}
 \usepackage[stable]{footmisc}
\usepackage{indentfirst}

\newcommand{\precaption}{\vspace{-0in}}
\newcommand{\postcaption}{\vspace{-0in}}
\newcommand{\presec}{\vspace{-0.0in}}
\newcommand{\postsec}{\vspace{-0.0in}}

\usepackage[linesnumbered,ruled]{algorithm2e}
\usepackage{xcolor}
\usepackage[font=small,labelfont=bf]{caption}

\newcommand{\sysnamepre}{{\sc $Pre$ }}

\newcommand{\sysnameupdten}{{\sc $FT$ }}

\newcommand{\sysnamepreNospace}{{\sc $Pre$}}

\newcommand{\sysnameupdtenNospace}{{\sc $FT$}}

\newcommand{\sysnamepremax}{{\sc $Pre$ }}

\newcommand{\sysnameupdtenmax}{{\sc $FT$ }}

\newcommand{\sysnamepremaxNospace}{{\sc $Pre$}}

\newcommand{\sysnameupdtenmaxNospace}{{\sc $FT$}}

\title{On the Robustness of Offensive Language Classifiers \thanks{~~This paper examines offensive language as a case study. The reader is cautioned that the paper contains unavoidable strong language given the nature of the research.}\thanks{~~Our code and data are available at: https://github.com/JonRusert/RobustnessOfOffensiveClassifiers}}

\author{Jonathan Rusert \\
  University of Iowa \\
  \texttt{jonathan-rusert}\\
  \texttt{@uiowa.edu} \\\And
  Zubair Shafiq \\
  University of California, Davis \\
  \texttt{zshafiq@ucdavis.edu} \\\And
  Padmini Srinivasan \\
  University of Iowa \\
  \texttt{padmini-srinivasan}\\
  \texttt{@uiowa.edu}
  }

\begin{document}

\maketitle

\begin{abstract}
%
Social media platforms are deploying machine learning based offensive language classification systems to combat hateful, racist, and other forms of offensive speech at scale. 
However, 
despite their real-world deployment, 
we do not yet comprehensively understand the extent to which offensive language classifiers are robust against adversarial attacks.
Prior work in this space is limited to studying robustness of offensive language classifiers against primitive attacks such as misspellings and extraneous spaces.
To address this gap, we systematically analyze the robustness of state-of-the-art offensive language classifiers against more crafty adversarial attacks that leverage greedy- and attention-based word selection and context-aware embeddings for word replacement.
Our results on multiple datasets show that these crafty adversarial attacks can degrade the accuracy of offensive language classifiers by more than 50\% while also being able to preserve the readability and meaning of the modified text.
%
\end{abstract}

\maketitle

\section{Introduction}

Online social media platforms are dealing with an unprecedented scale of offensive (e.g., hateful, threatening, profane, racist, and xenophobic) language   \cite{Hatefulc1:online,Communit4:online,Redditpolicy:online}.
Given the scale of the problem, online social media platforms now increasingly rely on  machine learning based systems to proactively and automatically detect offensive language \cite{Communit20:online,Anupdate61:online,TwitterRemoveAbuse:online,TwitterIncreaseAutomation:online}.
The research community is actively working to improve the quality of offensive language classification \cite{zampieri-etal-2020-semeval,zampieriOffensEval2019, liu-etal-2019-nuli, nikolov-radivchev-2019-nikolov, Mahata2019MIDASAS,Arango20hatespeechSIGIR,Agrawal18cyberECIR,Fortuna18hatespeechdetectionACMSurvey}.
A variety of offensive language classifiers ranging from traditional shallow models (SVM, Random Forest), deep learning models (CNN, LSTM, GRU), to transformer-based models (BERT, GPT-2) have been proposed in prior literature \cite{liu-etal-2019-nuli, nikolov-radivchev-2019-nikolov, Mahata2019MIDASAS}. 
Amongst these approaches, BERT-based transformer models have achieved  state-of-the-art performance while ensembles of deep learning models also generally perform well \cite{zampieriOffensEval2019,zampieri-etal-2020-semeval}.

It remains unclear whether the state-of-the-art offensive language classifiers are robust to adversarial attacks. 
While adversarial attacks are of broad interest in the ML/NLP community \cite{hsiehRobustness2019, behjati2019universal}, they are of particular interest for offensive language classification because malicious users can make subtle perturbations such that the offensive text is still intelligible to humans but evades detection by machine learning classifiers.
Prior work on the robustness of text classification is limited to analyzing the impact on classifiers of primitive adversarial changes such as deliberate misspellings \cite{Li19textbugger}, adding extraneous spaces \cite{grondahlLove2018}, or changing words with their synonyms \cite{jin2020bert, ren-etal-2019-generating, li-etal-2020-bert-attack}. 
However, the primitive attacks can be easily defended against---a spell checker can fix misspellings and a word segmenter can correctly identify word boundaries even with extra spaces \cite{Rojas-Galeano:2017:OOO:3079924.3032963,Li19textbugger}.
Additionally, a normal synonym substitution will not theoretically hold for offensive language as less offensive language will be substituted and thus meaning will be lost.
Crucially, we do not know how effective these text classifiers are against crafty adversarial attacks employing more advanced strategies for text modifications.

To address this gap, we analyze the robustness of offensive language classifiers against an adversary who uses a novel word embedding to identify word replacements and a surrogate offense classifier in a black-box setting to guide modifications.
This embedding is purpose-built to evade offensive language classifiers by leveraging an \textit{evasion  collection} that comprises of evasive offensive text gathered from online social media.
Using this embedding, the adversary modifies the offensive text while also being able to preserve text readability and semantics.
We present a comprehensive evaluation of the state-of-the-art BERT and CNN/LSTM based offensive language classifiers, as well as an offensive lexicon and Google's Perspective API, on two datasets.

We summarize our key contributions  below.

$\bullet$ We systematically study the ability of an adversary who uses a novel, crafty strategy to attack and bypass offensive language classifiers. 
    The adversary first builds a new embedding from a special evasion collection, then uses it alongside a surrogate offensive language classifier deployed in black-box mode to launch the attack.
    
$\bullet$ We explore variations of our adversarial strategy. These include greedy versus attention based selection of text words to replace. These also include two different versions of  embeddings for word substitutions.
    
$\bullet$ We evaluate robustness of state-of-the-art offensive language classifiers, as well as a real-world offensive language classification system on two datasets from Twitter and Reddit.
    Our results show that 50\% of our attacks cause an accuracy drop of $\geq$ 24\% and 69\% of attacks cause drops $\geq$ 20\% against classifiers across datasets.

\vspace{.05in} \noindent \textit{Ethics Statement:} We acknowledge that our research demonstrating attacks against offensive language classifiers could be used by bad agents. Our goal is to highlight the vulnerability within offensive language classifiers. We hope our work will inspire further research to improve their robustness against the presented and similar attacks. 

\presec
\section{Target Offensive Language Classifiers}
\label{sec:offenseModels}
\postsec


\subsection{Threat model}
%
The adversary's goal is to modify his/her offensive post in such a manner as to evade detection by offensive language classifiers while simultaneously preserving semantics and readability for humans. 
To make suitable modifications, the adversary is assumed to have black-box access to a surrogate offensive language classifier that is different from the one used by the online social media platform. 
The adversary leverages feedback from this surrogate classifier to guide modifications using a novel approach that we propose.
Our goal is to evaluate the extent to which the adversary can evade detection by an unknown offensive language classifier under this threat model.

\subsection{Offensive Language Classifiers}
\label{subsec: classifiers}
We evaluate the following offensive language classifiers under our threat model.

\begin{enumerate}

\item \textbf{NULI} \cite{liu-etal-2019-nuli} is a BERT \cite{BERT2018} based system trained on offensive language. During preprocessing, emojis are converted into English phrases\footnote{https://github.com/carpedm20/emoji} and hashtags are segmented\footnote{https://github.com/grantjenks/python-wordsegment}. This was the top-ranked system in OffensEval \cite{zampieriOffensEval2019}.


\item \textbf{Vradivchev} \cite{nikolov-radivchev-2019-nikolov} is also a BERT based system trained on offensive language data. The preprocessing step includes removing symbols ``@" and ``\#", tokenization and lowercasing, splitting hashtags, and removing stopwords. This was the second best system in OffensEval.

\item \textbf{MIDAS} \cite{Mahata2019MIDASAS} is a voting ensemble of three deep learning systems: a CNN, a BLSTM, and a BLSTM fed into a Bidirectional Gated Recurrent Unit (BGRU). This was the top non-BERT system in OffensEval\footnote{We implemented NULI, vradivchev and MIDAS with parameters reported in the cited papers. Our accuracies were within 1\% of the reported F1 score.}.

\item \textbf{Offensive Lexicon} \cite{wiegand-etal-2018-inducing} is a simple method that classifies a post as offensive if at least one word is in a lexicon of offensive words. 
We use their lexicon.

\item \textbf{Perspective API} \cite{Perspective} by 
Google (Jigsaw) provides a toxicity model that classifies whether a post is ``rude, disrespectful, or unreasonable.'' 
The production model uses a CNN trained with fine-tuned GloVe word embeddings and provides ``toxicity'' probability. 
We use 0.5 threshold to classify a post as offensive as in \citet{pavlopoulos2019convai}. 

\end{enumerate}

\presec
\section{Attack Methods} 
\label{sec:attacks}
\postsec
This section describes our adversarial attack method as well as a recent visual adversarial attack \cite{eger2019text} and a simpler attack \cite{grondahlLove2018} for baseline comparison.

\subsection{Proposed Attack} 
\label{sect:propattack}
The adversary's attack involves selecting words to replace in the input text and deciding on suitable replacements.

\vspace{.05in} \noindent
\textbf{Selection.} 
%
There are several ways to approach word selection for replacement.  Here we explore a greedy approach \cite{hsiehRobustness2019} 
and an approach 
using attention weights \cite{xu-etal-2018-unpaired}.
%
%
%

For the greedy approach, we first remove each word one at a time (retaining the rest in the text) and get the 
drop in classification probability for the text from the surrogate offensive classifier.
Words are removed until the offensive label is flipped (according to the classifier).
The removed words make up the full list of possible replacements.
The adversary then selects the word that causes the largest drop for replacement.
If replacing this word is insufficient to bypass the surrogate classifier then the word with the next largest drop is also selected for replacement and so on.

For the attention approach, we leverage a BLSTM with attention which is trained on the target classification task. 
Note that this BLSTM is different from the one found in MIDAS.
To select words, we give the input text to the BLSTM and examine the attention weights estimated during classification. 
The adversary selects the word with the highest attention weight. 
If replacing this word is insufficient to bypass the surrogate classifier then the word with the next largest attention weight is also selected for replacement and so on.

The attention approach can potentially find replacements that greedy approach may not.
Specifically, the greedy approach may miss instances where the combination of words cause offense rather than single words.
%


%

\begin{figure}
    \centering
    \includegraphics[width=1\columnwidth]{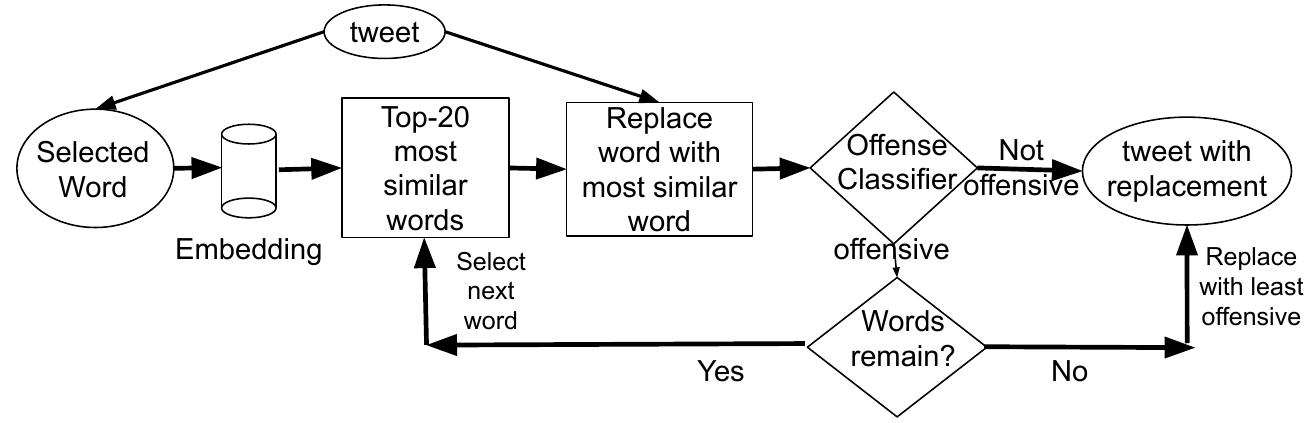}
    \precaption
    \vspace{-.3in}
    \caption{The design of our word replacement approach. First, select the  word's 20 most similar words as candidates. Next, replace the word with the most similar candidate and check text against surrogate classifier. 
    If this results in a not offensive classification, the process ends with this word as replacement. Otherwise, continue the process with the next most similar candidate. If no candidates remain, choose the one which causes the greatest drop in classification probability.}
    \postcaption
    \label{fig:replacement}
\vspace{-.1in}
\end{figure}

\vspace{.05in} \noindent
\textbf{Replacement.}  
Figure \ref{fig:replacement} depicts our framework for substituting the selected word with another word. 
First, a candidate list of 20 most similar words (closest vectors) is obtained from an embedding space.
%
%
Next, we replace the selected word with its most similar word and check the modified text against the surrogate classifier. 
If the modified text is declared not offensive, then this word is chosen as the replacement. 
Otherwise, the process continues with the next most similar word. 
If the candidate list is exhausted without misclassification by the surrogate classifier, we choose the  replacement word which causes the largest drop in classification probability.


\vspace{.05in} \noindent
\textbf{Embeddings.}
The key idea here is to design a context-aware word embedding for crafty replacements.
To this end, we first build a text collection of 13 million deleted tweets through retrospective analysis using the Twitter API \cite{thomas11retro,le19suspended}.
Next we filter out the tweets from this set that are labeled as offensive by any of the offensive language classifiers in Section \ref{subsec: classifiers}.\footnote{Note that Perspective did not participate in this tagging due to query limits of the API.}
The remaining set of 8.5 million deleted tweets contains offensive tweets that were likely flagged by users or human moderators.\footnote{It is possible that some of these tweets were deleted for other reasons that are unknown to us.}
We expect this set of deleted tweets to contain crafty substitutions and expressions that are likely to evade detection by state-of-the-art offensive language classifiers.
We refer to this set of deleted tweets as the \emph{evasion} collection and this is the data that the adversary uses to train word embeddings.
We explore the following embeddings:

\begin{figure}
    \centering
    \includegraphics[width=1\columnwidth]{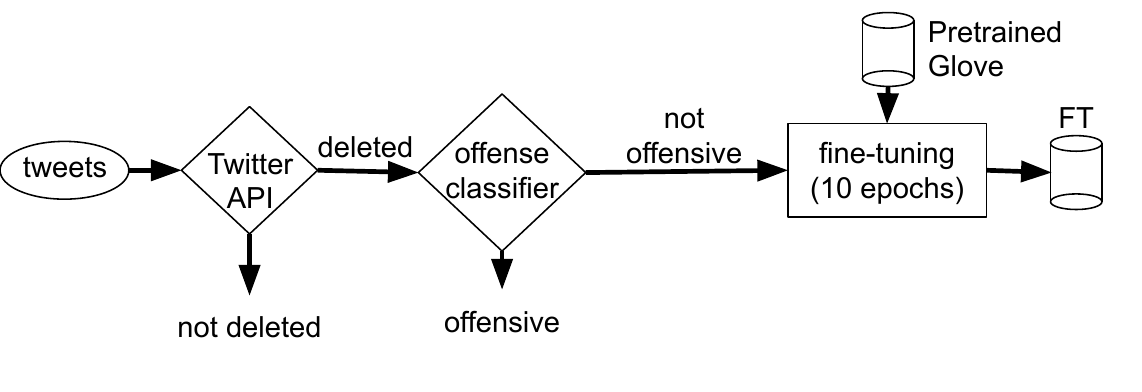}
    \precaption
    \vspace{-.3in}
    \caption{Our approach for creating \sysnameupdten. First, 13 million deleted tweets are identified via retrospective analysis using the Twitter API \cite{le19suspended,thomas11retro}. Next, the tweets are checked against offensive language classifiers to remove those detected as offensive. Finally, for \sysnameupdtenNospace, fine-tune the pretrained (\sysnamepreNospace) with the remaining tweets for 10 epochs.}
    \postcaption
    \label{fig:FT}
    \vspace{-.1in}
\end{figure}


\begin{enumerate}

\item GloVe pretrained Twitter embedding (\sysnamepreNospace): These are pretrained GloVe embeddings on 2 billion tweets. The vocabulary size of this model is 1,193,514 tokens. This represents a baseline off-the-shelf word embedding.


\item GloVe embedding fine-tuned with evasion collection (\sysnameupdtenNospace):  We use the evasion collection to fine-tune the pretrained GloVe  embeddings. Fine tuning is done over 10 epochs. The resulting vocabulary size is 1,312,106 tokens. Figure \ref{fig:FT} illustrates this approach.

\end{enumerate}

\vspace{.05in} \noindent
\textbf{Insights into the embeddings.} 
Our intuition of crafty substitutions being present in the evasion collection is backed up by examination of the embeddings. 
Using a set of offensive words as probes we find that on average the  position of the first evasive
word amongst the 20 most similar words in \sysnamepre 
is 11, while for \sysnameupdten this number is 3, implying that \sysnameupdten is more likely to offer an evasive replacement. 
We expand on these insights and analysis in Section \ref{sec:discussion}.
Furthermore, as fine-tuning the embeddings may introduce garbage words (non english, often meaningless words) as replacements, we add in a filter to the candidates when using the FT embeddings. This filter only allows candidates which have been used in tweets by 3 distinct authors in the \textit{evasion} dataset. Finally, as checking every candidate can be time consuming and inefficient, we apply this filter only when we substitute text words that were not in the original \textit{Pre} embeddings.

\subsection{Other Attacks}

\noindent \textbf{VIPER.} We implement a recent visual adversarial attack called VIPER \cite{eger2019text} that aims to generate adversarial text for any classification task. 
VIPER (VIsual PERturber) replaces characters in the text with visually nearest neighbors determined from a visual embedding space. 
Each character present in the text is selected for replacement with a fixed probability $p$. 
VIPER strategically chooses replacements from non standard unicode characters assuming that systems rarely train outside the standard unicode space.
As the main comparison, we choose their description-based character embedding space (DCES) in our experiments since it had the best tradeoff between attack success and readability. 
DCES represents characters by their unicode textual descriptions. 
The nearest neighbor substitute is the character whose description refers to the same letter in the same case.
We also compare with their simpler, easy character embedding space (ECES), which contains only nearest neighbor for character replacement.
We used VIPER with $p$ = 0.1 and 0.4, the first for better readability and the second for better likelihood of attack success.
Note that higher $p$ values correspond to more changes in the text.

\noindent \textbf{Grondahl.}
\citet{grondahlLove2018} explored rather simple attack methods such as modifying whitespace and misdirection by adding a misleading word. 
We implement several of their adversarial attacks.
These are: adding a space after every character, removing all spaces between characters, adding the word `love' to the input text, and finally removing all spaces then adding `love.'
This last attack strategy outperformed others in their evaluation.

\section{Experimental Setup}
\label{sec:experimentalsetup}

\subsection{Datasets}\label{sect:dataset}
%

\vspace{.05in} \noindent
\textbf{Offensive Language Identification Dataset (OLID).}
OLID was used in SemEval-6 2019: OffensEval, a shared task on classifying offensive language \cite{zampieri-etal-2019-predicting}. 
This collection is annotated by experienced annotators to ensure high quality.
OLID contains 14,100 English tweets (text only): split into 13,240 (4,400 offensive, 8,840 non-offensive) training tweets and 860 (240 offensive, 620 non-offensive) test tweets. 

\noindent
\textbf{Semi-Supervised Offensive Language
Identification Dataset (SOLID)}. SOLID is an expansion of OLID used in SemEval 2020: OffensEval, which continued the task of classifying offensive language \cite{rosenthal2020large}. 
SOLID was constructed from tweets via semi-supervised manner using democratic co-training with OLID as a seed dataset. SOLID contains 9,000,000 tweets as an expansion for training, and 5,993 test tweets, (3,002 offensive, 2,991 non-offensive).


\subsection{Evaluation Metrics} \label{sect:evals}

\vspace{.05in} \noindent
\textbf{Drop in Accuracy:}

\(
    \Delta = \text{Accuracy}_{\text{Original}} - \text{Accuracy}_{\text{Modified}},
\) 

where $\text{Accuracy}_{\text{Original}}$ is the classifier's accuracy on original text and $\text{Accuracy}_{\text{Modified}}$ is the classifier's accuracy on the modified text.
Larger drops imply better evasion of offensive language classifiers by the adversary.

\vspace{.05in} \noindent
\textbf{Readability and semantic preservation:}  
We measure readability of the modified text and its semantic preservation through manual evaluation.
More specifically, for readability, human reviewers are asked to examine the modified text and rate it as one of: \{`The text is easy to read', `The text can be read with some difficulty', `The text is hard to read'\}. 
For semantic preservation, reviewers are given the original texts alongside the modified versions and are asked whether `text B' (modified text) conveys the same meaning as `text A' (original text). 
The choices are \{`Yes, Text B conveys the same meaning as Text A', `Text B conveys partially the meaning of Text A', `No, Text B does not convey the same meaning as Text A'\}.

\subsection{Experiment Design}
We use the OLID and SOLID test sets to assess the success of our attack strategies. 
Amongst the several offensive language classifiers considered in this work (see Section \ref{subsec: classifiers}), we make one classifier available to the adversary as a surrogate black-box classifier to guide adversarial modification of each test tweet.
Note that we do not use Lexicon as an internal classifier as it does not provide useful feedback (only returning 0 or 1 for positive class probabilities).
We then evaluate the drop in classification accuracy ($\Delta$) for each of the remaining classifiers.
%
%

\presec
\section{Results}
\label{sec:results}
\postsec
In this section, we first present the results of our proposed adversarial attack approach and then those of existing approaches from prior literature on the OLID dataset. Evaluation was also performed on the SOLID dataset and the results followed a similar trend. Full results for all attacks are located in the appendix.

\subsection{Our adversarial attack}\label{sect:accuracy}

Table \ref{tab:OFFResultsMAX} presents the results on the OLID dataset. 
Rows specify the attack strategy.
The first column identifies the surrogate offensive language classifier used by the adversary to guide modifications. 
%
%
The remaining columns specify the offensive language classifier whose robustness is being evaluated. 
Cell values are  drops in accuracy after adversarial modification.
Accuracy here refers to the percentage of offensive tweets correctly predicted as offensive.
%
%
Classification accuracy for original text is given in the first row of the table.
So for example, the final accuracy for NULI where the adversary uses GS-Pre and MIDAS is 44 (61-17).
Blocks of rows labeled with prefix GS stand for results with greedy word selection strategy while AS stand for results with BLSTM-attention based word selection.
Note that diagonal entries, where the surrogate classifier is the same as the one being tested for robustness are ignored because the adversary is expected to be quite successful under this condition.
We indeed find that the accuracy drops close to 0\% in these cases.
Additionally, for the Lexicon based method, we find it does not perform as well as the other classifiers, thereby excluding it from the state-of-the-art (SOTA) category.

\begin{table*}[]
    \footnotesize
    \centering
    \begin{tabular}{c|c||ccccc|c}
         &  & \multicolumn{5}{c}{Drop in Classification Accuracy} & \\\hline
          &  & NULI & vradivchev & MIDAS & Perspective & Lexicon & Avg. Drop\\\hline
          
          \multicolumn{2}{c}{No Attack Accuracy \%} & 61 & 69 & 66 & 68 & 54\\\hline\hline
          & Surrogate Classifier & & & & & & \\\hline
         \parbox[t]{2mm}{\multirow{5}{*}{\rotatebox[origin=c]{90}{GS - \sysnamepremax}}} & NULI & - & 41 & 33 & 34 & 24 & \textit{33}\\
         & vradivchev & 28 & - & 33 & 28 & 22 & 28\\
         & MIDAS & 17 & 35 & - & 26 & 19 & 24\\
         & Perspective & 20 & 36 & 30 & - & 17 & 26\\ \cline{2-8}
          & Average Drop & 22 & \textbf{37} & \textbf{32} & \textbf{29} & \textbf{21} & \\
         \hline \hline
         \parbox[t]{2mm}{\multirow{5}{*}{\rotatebox[origin=c]{90}{GS - \sysnameupdtenmax}}} & NULI &
         - & 46 & 30 & 31 & 19 & \textit{32}\\
         & vradivchev & 39 & - & 30 & 26 & 18 & 28\\
         & MIDAS & 18 & 29 & - & 23 & 13 & 21\\
         & Perspective & 22 & 37 & 28 & - & 13 & 25\\\cline{2-8}
           & Average  Drop &  \textbf{26} & \textbf{37} & 29 & 27 & 16 & \textbf{\textit{}}\\
         \hline\hline
 
           \parbox[t]{2mm}{\multirow{5}{*}{\rotatebox[origin=c]{90}{AS - \sysnamepremax}}} & NULI & - & 36 & 19 & 19  & 15 & 22\\
           & vradivchev & 22 & - & 18  & 19 & 17 & 19\\
           & MIDAS & 13 & 34 & - & 20 & 15 & 21\\
           & Perspective & 17 & 37 & 23 & - & 16 & \textit{23}\\\cline{2-8}
           & Average Drop & 17  & 36  & 20 & 19  & 16 &   \\\hline\hline

           \parbox[t]{2mm}{\multirow{5}{*}{\rotatebox[origin=c]{90}{AS - \sysnameupdtenmax}}} & NULI & - & 39 & 18 & 17 & 15 & 22 \\
           & vradivchev & 23 & - & 17 & 15 & 15 & 18\\
           & MIDAS & 11 & 27 & - & 17 & 12 & 17\\
           & Perspective & 17 & 40 & 21 & - & 16 & \textit{24}\\\cline{2-8}
           & Average Drop & 17 & 35 & 19 & 16 & 15 &  \\\hline\hline


          \end{tabular}
    \caption{Robustness results on OLID  with our attack model. Columns show accuracy drop. The approach is specified as \textit{selection - replacement} where \textit{selection = \{Greedy Select (GS), Attention Select (AS)\}} and \textit{replacement = \{\sysnamepremaxNospace, \sysnameupdtenmaxNospace\}}. 
    Note that the BLSTM used for $AS$ can be used as an internal classifier but performed poorly so was not included. 
    The adversarial, surrogate classifier is indicated in column 1.  The first row presents baseline classification accuracies (\%) before attacks. Therefore the resulting accuracies can be calculated by subtracting the drop from the original accuracy.}
    \postcaption
    \label{tab:OFFResultsMAX}
\end{table*}


\vspace{.05in} \noindent
\textbf{Offensive language classifiers are susceptible to our adversarial attacks.}
Table \ref{tab:OFFResultsMAX}  shows that our adversarial attacks are quite successful against crafty offensive language classifiers.
For OLID, classifiers see a drop of accuracy in the range of 11--46\footnote{Note: The BLSTM attention classifier used for attention based word selection could also be used as an internal classifier. However, since this strategy did not perform as well as SOTA classifiers so we do not include these results in the main analysis.}.
In fact, ~50\% of attacks cause a drop of $\geq$ 24 and 69\% of attacks cause a drop of $\geq$ 20.  
This shows the vulnerability of offensive language classifiers and their vulnerability under our threat model.

\vspace{.05in} \noindent
\textbf{Greedy select (GS) outperforms Attention select (AS) attacks.}
Greedy Select achieves higher average drops in accuracy across classifiers. 
For example, GS - \sysnameupdtenmax achieves an average drop of 26 against NULI while AS - \sysnameupdtenmax achieves only a drop of 17.
This holds true for both replacement embeddings.
Although lower, AS still achieves strong drops against vradivchev (average of 35). 
This indicates the strength of a greedy approach, however, attention selection may be more viable in a setting where the number of queries is limited.

\vspace{.05in} \noindent
\textbf{\textit{FT} embeddings and \textit{Pre} embeddings see success against different systems.}  
Comparing to \sysnamepremax embeddings, \sysnameupdtenmax we see different leads in dropped accuracy depending on the classifier. \sysnameupdtenmax see great success against NULI and vradivchev, while \sysnamepremax see success against the other three. This indicates that the \textit{evasion} collection can help add power, especially against popular (BERT-based) classifiers.

\vspace{.05in} \noindent
\textbf{NULI and vradivchev, BERT based classifiers, are the most and least robust to attacks.} Focusing on the GS - \sysnameupdtenmax embedding, NULI has a mean drop in accuracy of 26 (range: 18 - 39), the lowest across SOTA offensive language classifiers.
%
%
In contrast, vradivchev, performs the best with accuracy of 69 but is also the most vulnerable to our attack model with a mean drop in accuracy of 37 (range: 29 - 46), the highest drop of any offensive language classifier.
This mean is 27 for Perspective and 29 for MIDAS.
The stark difference between the two BERT systems' robustness most likely stems from the preprocessing step.
BERT is a context-aware system. While NULI's preprocessing helps add context (e.g. converting emojis to text), vradivchev's hinders it. Specifically, vradivchev removes stop words. This could be a problem as removing this additional information causes the system to miss out on context during training. Then, as the attack is more likely to focus on changing non-stop words, vradivchev then loses both contextual information (via stop word removal) as well as offense indicating tokens (the main information it focused on during training).

\vspace{.05in} \noindent
\textbf{NULI is the most effective surrogate classifier for the adversary while MIDAS the least effective.} 
Again, focusing on GS-\sysnameupdtenmax, NULI helps the adversary as the surrogate classifier the most by causing an average accuracy drop of 32 (range: 19 - 46), compared to vradivchev (avg: 28, range: 18 - 39), Perspective (avg: 25, range: 13 - 37), and MIDAS (avg: 21, range: 13 - 29).
This again emphasizes BERT based methods' ability to understand context and use it effectively in attacks, also seen in previous research \cite{li-etal-2020-bert-attack}.

\noindent \textbf{Replication of Results.} We replicate our results on a Reddit dataset in the appendix. 

%
%
%
%

\subsection{Other attacks}
\noindent \textbf{Grondahl.}
%
Table \ref{tab:charPre} shows the results when methods proposed by \citet{grondahlLove2018} are used to obfuscate.
Note that this approach does not use a surrogate classifier.
The simpler whitespace and `love' word based attacks proposed by \citet{grondahlLove2018} have little to no effect on offensive language classifiers which contain a word segmentation pre-processing step.
These classifiers include NULI (average drop: -3), vradivchev (average drop: 11), and Lexicon (average drop:0).
MIDAS being ill equipped in this regard sees a drop of 64 when all spaces are removed and `love' is added to the text.
However, when we add a simple word segmentation step during pre-processing the attack loses effectiveness. For example, the ``Remove space, Add `love''' attack is reduced to a drop of 33 with this shielding pre-processing step, compared to 64 without it.
Similarly, Perspective also sees drops up to 38 in these settings. 
%


%

\noindent\textbf{VIPER.}
%
Like a whitespace attack, VIPER attacks can be easily prevented using a trivial text pre-processing step. 
To demonstrate this, we added a pre-processing `shielding' step to each system which replaces any non-standard characters with standard ones.
The results for shielded VIPER attacks are found in Table \ref{tab:charPre} (Note: The full results for non-shielding against VIPER are found in the appendix.).
This is in essence logically the reverse  of VIPER's obfuscation by character translation process.
Non-standard characters are those which exist outside `a-zA-Z', numbers, and punctuation.
To do this, as do the VIPER authors, we leverage the NamesList from the unicode database\footnote{\url{https://www.unicode.org/Public/UCD/latest/ucd/NamesList.txt}}. 
For any non-standard character, the description is searched for in the NamesList and the character which appears after ``LETTER'' in the description is used for substitution. 
For example, '$\hat{a}$'is described as ``LATIN SMALL LETTER A WITH INVERTED BREVE'', and hence would be replaced with `a'. 
This simple pre-processing step reduces VIPER's average attack rate from 37 to 7 as shown in the VIPER results. 
In contrast, our proposed attack is not  preventable through such simple pre-processing. 
%

\begin{table*}[]
    \small
    \centering
    \begin{tabular}{cc||cccccc}
         & &  \multicolumn{5}{c}{Drop in Classification Accuracy}\\\hline
         & & NULI & vradivchev & MIDAS & Perspective & Lexicon & Avg. Drop \\\hline
         
         \multicolumn{2}{c}{No Attack Accuracy \%} & 61 & 69 & 66 & 68 & 54 & \\\hline\hline
         
         \parbox[t]{2mm}{\multirow{4}{*}{\rotatebox[origin=c]{90}{Grondahl}}} & Add Space & -6 & 8 & 51 & 2 & 0 & 11 \\
         & Remove Space & -6 & 8 & 66 & 34 & 0 & 20\\
         & Add `love' & 0 & 14 & 8 & 1 & 0 & 3\\
          & Remove Space, Add `love' & 0 & 14 & 64 & 38 & 0 & 23 \\\hline\hline
         
          \multicolumn{7}{c}{Drop in Classification Accuracy after Shielding using Character Preprocessing}\\\hline
          \multicolumn{2}{c||}{VIPER(0.1, DCES)}   & -5 & 12 & 17 & 3 & 3 & 6\\  
          \multicolumn{2}{c||}{VIPER(0.4, DCES)}  & 5 & 20 & 23 & 8 & 10 & 13\\
          \multicolumn{2}{c||}{VIPER(0.1, ECES)} & -7 & 8 & 18 & 1 & 0 & 4\\
          \multicolumn{2}{c||}{VIPER(0.4, ECES)} & -7 & 8 & 18 & 1 & 0 & 4\\\hline\hline
          
          \end{tabular}
    \caption{Robustness results on OLID  against Grondahl and VIPER attacks (with and without shielding with simple character replacement pre-processing step). Columns show accuracy drop. The first row presents classification accuracies before attacks.}
    \postcaption
    \label{tab:charPre}
\end{table*}

\subsection{SOLID Results}
\begin{table*}[]
    \small
    \centering
    \begin{tabular}{c|c||ccccc|c}
         &  & \multicolumn{5}{c}{Drop in Classification Accuracy} & \\\hline
          &  & NULI & vradivchev & MIDAS & Perspective & Lexicon & Avg. Drop\\\hline
          
          \multicolumn{2}{c}{No Attack Accuracy \%} & 96 & 93 & 99 & 97 & 82 \\\hline\hline
          & Surrogate Classifier & & & & & & \\\hline

         \parbox[t]{2mm}{\multirow{5}{*}{\rotatebox[origin=c]{90}{GS - \sysnameupdtenmax}}} & NULI &
         - & 69 & 45 & 50 & 49 & 53\\
         & vradivchev & 54 & - & 42 & 40 & 43 & 41\\
         & MIDAS & 33 & 38 & - & 39 & 34 & 37\\
         & Perspective & 46 & 54 & 43 & - & 40 & 46\\\cline{2-8}
           & Average  Drop & 44 & 54 & 43 & 43 & 41 &  \\
         \hline\hline

          \end{tabular}
    \caption{Robustness results on SOLID  with our attack model. Columns show accuracy drop. The adversarial, surrogate classifier is indicated in column 1.  The first row presents baseline classification accuracies (\%) before attacks. Therefore the resulting accuracies can be calculated by subtracting the drop from the original accuracy.}
    \postcaption
    \label{tab:SOLIDResults}
\end{table*}

The attack results against SOLID are found in Table \ref{tab:SOLIDResults}. We see similar attack success as seen in OLID, finding even greater drops. Specifically, 75\% of attacks cause a drop of $\geq$ 40 and 100\% attacks cause a drop of at least 33.

\subsection{Quality of Adversarial Examples}

\begin{table}[]
    \small
    \centering
    \begin{tabular}{c|ccc}
         Adversarial & \multicolumn{3}{c}{Readability}  \\
         Attack & Yes & Partially & No \\ \hline
          \sysnameupdtenmax & 35 & 13 & 2\\
          $[\%]$ & 70.0 & 26.0 & 4.0\\\hline
          Original & 37 & 13 & 0\\
          $[\%]$ & 74.0 & 26.0 & 0.0\\

         \hline\hline
         & \multicolumn{3}{c}{Conveys same meaning}\\
         & Yes & Partially & No \\\hline
         \sysnameupdtenmax & 31 & 17 & 2\\
         $[\%]$ & 62.0 & 34.0 & 2.0
         \\\hline
         
    \end{tabular}
    \caption{Results of human readability and meaning comparison assessments. Majority voting was used to combine all three annotators' answers into one vote. }
    \label{tab:readability}
\end{table}

\vspace{.05in} \noindent
\textbf{\sysnameupdtenmax embeddings maintain a majority of the meaning and readability.}
We test readability of a sample of 50 tweets from the SOLID dataset, of which all were modified by \sysnameupdtenmax.
%
%
We asked three crowdworkers to assess the 50 tweets for readability.
For comparison, we asked additional crowdworkers to assess the readability of the original texts. This helps explore the true drop in readability of a text.
Additionally, we showed three other crowdworkers the original text as well and asked them to assess if the obfuscated texted conveyed the same meaning as the original (see Section \ref{sect:evals} for details).
We finally combined the crowdworkers votes by taking a majority vote for each example.

Table \ref{tab:readability} presents the results. 
We find that \sysnameupdten scored slightly less in terms of readability than the original texts, but finds replacements with similar meaning.
Specifically, readability drops from 74\% to 70\% for fully readable, but
nearly two thirds retain the same meaning and 96\% retain at least partial meaning.
These numbers help indicate the strength of the attack, even when leveraging a crafty collection of word substitutions.
%
%

%

%
%
To provide insights into texts which retained full meaning versus partial, Table \ref{tab:readabilityExamples} shows a few examples of tweets in their respective categories as voted by crowdworkers.
\sysnameupdtenmax is able to find many appropriate, non traditional replacements. 
For example, ``shit'' is replaced with ``shxt'', in several instances which helps maintain meaning while evading classification. 
As another example, ``phising'' a mispelling of ``phishing'' is substituted for ``fake''. 
In context, this substitution makes sense.
Note that while some examples are misspellings, these crafty modifications are ones that are mined from our large evasion text collection and not algorithmically-generated mispellings. 
However, some errors are found after replacement. For example, in the \textit{Not Similar} instance \sysnameupdtenmax replaces ``fuck'' with ``bruh'', and ``shut'' with ``walked''.
These errors demonstrate room for improvement when selecting a candidate.
%



\begin{table}[!t]
    \small
    \centering
    \begin{tabular}{|p{3.5cm}|p{3.5cm}|}
    \hline
         Original Tweet &  \sysnameupdtenmax \\\hline 
         \multicolumn{2}{|c|}{Very Similar}
        \\\hline
         The amount of fake pages that watches my snap is insane I hate bitches so bad g 
        & The amount of phising pages that watches my snap is unbelievable I hate bitchs so bad g 
            \\\hline
        @USER Nigga is a rapist, he deserves all the shitty things that happen to him & @USER Nigha is a peado, he deserves all the awful things that happen to him \\\hline
        arab men are some of the most egotistical, disrespectful, pieces of shit you’ll ever meet & arab men are some of the most egotistical, inconsiderate, pieces of shxt you’will ever meet \\\hline
    
        \multicolumn{2}{|c|}{Somewhat Similar}
        \\\hline
        @USER Look at his lame ass with that shit eating smile... he is so vile. & @USER Look at his lame asf with that shxt eating smile... he is so discusting. \\\hline
        Yes I am a triple threat. A bad bitch, A dumb bitch, And a sensitive bitch & Yes I am a triple approach. A bad hoe, A rude hoe, And a hostile niggah \\\hline
        @USER You're a shameless pig but you knew that already.  Just a reminder.  Enjoy your jail time. & @USER You're a self-promotion baboon but you knew that already. Just a reminder. Enjoy your jail time. \\\hline
        
         \multicolumn{2}{|c|}{Not Similar}
        \\\hline
        shut the fuck omg no one cares damn & walked the bruh omg no one cares lmaoo
        \\\hline

             \end{tabular}
    \caption{Examples of tweets in majority voted categories from crowdworkers. }
    \postcaption
    \label{tab:readabilityExamples}
\end{table}

\presec
\section{Analysis of embeddings} \vspace{-.1in}
\label{sec:discussion}
\postsec

As discussed in Section \ref{sect:propattack},
the adversary's strategy is to make crafty word replacements using a new embedding generated from an evasion collection (here made of deleted tweets not detected by an offense classifier). 
Results show that these embeddings successfully support the adversary at evading offensive language classifiers while maintaining readability and semantics. 
For further insights, we compare the off-the-shelf pretrained (\sysnamepreNospace) embedding with the embedding fine-tuned on the evasion collection (\sysnameupdtenNospace). 
We examine the embeddings using the 59  words as probes which are both in the offensive Lexicon \cite{wiegand-etal-2018-inducing} and in the OLID test.
%
%
%
For each word we get the 20 most similar words from \sysnamepre and from \sysnameupdten for comparison.

\vspace{.05in} \noindent \textbf{Fine-tuned embeddings move evasive substitute words closer to offensive probe words.} 
We calculate the average position of the first evasive word\footnote{Evasion is determined by Perspective API.} amongst the 20 most similar words. 
\sysnamepre has an average distance of 11, while \sysnameupdten has an average distance of 3.
Thus, on average, \sysnameupdten is more likely to find evasive replacements. 
For example, in \sysnamepre \textit{dispicable} appears as the 3rd most similar word to \textit{despicable}, but it is the most similar in \sysnameupdtenNospace.
Since \sysnameupdten could contain some unintelligible words, we repeat the experiment to filter out substitute words used by less than 3 different users. 
The same overall trend still holds.

\vspace{.05in} \noindent \textbf{Updated embeddings learn creative replacements.}
We manually compare the entries in the two lists (\sysnameupdten and \sysnamepreNospace) of substitute words for each probe word. 
\sysnameupdten learns creative replacements absent in \sysnamepreNospace.
Examples include the word \textit{azz} being the most similar word to \textit{ass} in \sysnameupdtenNospace, but being absent within the most similar word list for \sysnamepreNospace. 
Similarly, \textit{niggah} appears as a replacement for \textit{bitch} in \sysnameupdtenNospace, but not in \sysnamepreNospace. 
These examples, along with the previous distance analysis, illustrate the craftiness in our evasion dataset.

\presec
\section{Related Work}
\label{sec:related}
\postsec
We first review related work on robustness of text classification in general and then closely related research on evading offensive language classifiers.

\vspace{.05in} \noindent
\textbf{Evading Text Classifiers.}
Prior work has explored ways to evade text classification in general. 
\citet{Li19textbugger} showed that  character-level perturbations such as misspellings and word-level perturbations using off-the-shelf GloVe embeddings can evade text classifiers. 
\citet{deri-knight-2015-make} proposed an approach to create portmanteaus, which could be extended to adversarial texts. 
\citet{behjati2019universal} added a sequence of words to any input to evade text classifiers.
\citet{zhaoAdversarial2017} proposed a GAN to generate adversarial attacks on text classification tasks.
\cite{li-etal-2020-bert-attack} leverage BERT to propose solutions for replacement words, \cite{jin2020bert} leverage word embeddings, and \cite{ren-etal-2019-generating} leverage WordNet.
In contrast to prior work evading text classifiers, our work includes approaches to leverage embeddings built from a special evasion text collection. 

\vspace{.05in} \noindent
\textbf{Robustness of Text Classifiers.}
Our work is also relevant to prior studies of the robustness of text classifiers to adversarial inputs. 
\citet{Rojas-Galeano:2017:OOO:3079924.3032963} showed that primitive adversarial attacks (e.g., misspellings) can be detected and countered using edit distance.
\citet{hsiehRobustness2019} evaluated the robustness of self-attentive models in tasks of sentiment analysis, machine translation, and textual entailment.
%
%
We examine robustness of similar models, however, we fine tune our embeddings to be task specific, while they do not, and we also test on the state-of-the-art offensive language classifiers.

\vspace{.05in} \noindent
\textbf{Evading Offensive Language Classifiers.}
\citet{grondahlLove2018} examined robustness of hate speech classifiers against adding typos, whitespace, and non-hate words to text.  
As discussed earlier, prior work has shown that such primitive perturbations can be detected and reversed \cite{Li19textbugger,Rojas-Galeano:2017:OOO:3079924.3032963}.
In contrast, we focus on more crafty text perturbations in our work. 
\citet{ji-knight-2018-creative} surveyed the ways text has been encoded by humans to avoid censorship and explain challenges which automated systems would have to overcome.
This work does not propose an automated approach for text perturbation.  
%
%
\citet{eger2019text} proposed VIPER for visual adversarial attacks. 
We implemented VIPER and \cite{grondahlLove2018} as baseline attacks and showed that our approach is more successful overall.
Overall, our work advances the research in this space by investigating robustness of offensive language classifiers against crafty adversarial attacks.


%

\presec
\section{Conclusion}
\postsec
\vspace{-.1in}
In this paper, we showed that state-of-the-art offensive language classifiers are vulnerable to crafty adversarial attacks.
Our proposed adversarial attacks that leverage greedy  and attention-based word selection and context-aware embeddings for word replacement were able to evade offensive language classifiers while preserving readability and semantics much better than prior simpler adversarial attacks.
We report accuracy drops of up to 46 points or 67\% against state-of-the-art offensive language classifiers. 
Furthermore, unlike VIPER and simpler attacks, our proposed attack cannot be easily prevented using pre-processing strategies. 
The user study showed that our adversarial attack was able to maintain similar readability with only a slight drop in semantic preservation.

%


Our work also suggests ways to improve the robustness of offensive language classifiers through adversarial training \cite{DBLP:conf/iclr/KurakinGB17, DBLP:conf/iclr/MadryMSTV18, DBLP:conf/iclr/TramerKPGBM18}.
More specifically, our attack relies on the \textit{evasion} collection, which contains crafty adversarial examples that evade detection by offensive language classifiers but are flagged based on manual feedback by users or human moderators.
Thus, offensive language classifiers can be adversarially trained on the latest \textit{evasion} collection from time to time to improve their robustness to the ever evolving adversarial attacks. 
In this context it is noteworthy that continuous availability of large-scale manual feedback is quite unique to the problem of offensive language classification, where popular online social media platforms employ thousands of human moderators \cite{NYUConte91:online}.

\bibliographystyle{acl_natbib}
\bibliography{main}

\appendix
\newpage
\onecolumn
\newpage
\twocolumn
\section{Full Results}
Table \ref{tab:allResults} shows full results for all attacks, including our attack, Grondahl ,and VIPER attacks before and after shielding. 

\begin{table*}[]
    \footnotesize
    \centering
    \begin{tabular}{c|c||ccccc|c}
         &  & \multicolumn{5}{c}{Drop in Classification Accuracy} & \\\hline
          &  & NULI & vradivchev & MIDAS & Perspective & Lexicon & Avg. Drop\\\hline
          
          \multicolumn{2}{c}{No Attack Accuracy \%} & 61 & 69 & 66 & 68 & 54\\\hline\hline
          & Surrogate Classifier & & & & & & \\\hline
         \parbox[t]{2mm}{\multirow{5}{*}{\rotatebox[origin=c]{90}{GS - \sysnamepremax}}} & NULI & - & 41 & 33 & 34 & 24 & \textit{33}\\
         & vradivchev & 28 & - & 33 & 28 & 22 & 28\\
         & MIDAS & 17 & 35 & - & 26 & 19 & 24\\
         & Perspective & 20 & 36 & 30 & - & 17 & 26\\ \cline{2-8}
          & Average Drop & 22 & \textbf{37} & \textbf{32} & \textbf{29} & \textbf{21} & \\
         \hline \hline
         \parbox[t]{2mm}{\multirow{5}{*}{\rotatebox[origin=c]{90}{GS - \sysnameupdtenmax}}} & NULI &
         - & 46 & 30 & 31 & 19 & \textit{32}\\
         & vradivchev & 39 & - & 30 & 26 & 18 & 28\\
         & MIDAS & 18 & 29 & - & 23 & 13 & 21\\
         & Perspective & 22 & 37 & 28 & - & 13 & 25\\\cline{2-8}
           & Average  Drop &  \textbf{26} & \textbf{37} & 29 & 27 & 16 & \textbf{\textit{}}\\
         \hline\hline
 
           \parbox[t]{2mm}{\multirow{5}{*}{\rotatebox[origin=c]{90}{AS - \sysnamepremax}}} & NULI & - & 36 & 19 & 19  & 15 & 22\\
           & vradivchev & 22 & - & 18  & 19 & 17 & 19\\
           & MIDAS & 13 & 34 & - & 20 & 15 & 21\\
           & Perspective & 17 & 37 & 23 & - & 16 & \textit{23}\\\cline{2-8}
           & Average Drop & 17  & 36  & 20 & 19  & 16 &   \\\hline\hline

           \parbox[t]{2mm}{\multirow{5}{*}{\rotatebox[origin=c]{90}{AS - \sysnameupdtenmax}}} & NULI & - & 39 & 18 & 17 & 15 & 22 \\
           & vradivchev & 23 & - & 17 & 15 & 15 & 18\\
           & MIDAS & 11 & 27 & - & 17 & 12 & 17\\
           & Perspective & 17 & 40 & 21 & - & 16 & \textit{24}\\\cline{2-8}
           & Average Drop & 17 & 35 & 19 & 16 & 15 &  \\\hline\hline


         \multicolumn{2}{c||}{VIPER(0.1, DCES)}   & 16 & 30 & 19 & 16 & 20 & 20\\  
          \multicolumn{2}{c||}{VIPER(0.4, DCES)}  & 55 & 66 & 58 &  54 & 39 & 54\\
          \multicolumn{2}{c||}{VIPER(0.1, ECES)} & 20 & 29 & 21 & 15 & 18 & 21\\
          \multicolumn{2}{c||}{VIPER(0.4, ECES)} & 54 & 63 & 57 & 44 & 48 & 53\\\hline\hline
          \multicolumn{7}{c}{Drop in Classification Accuracy after Shielding using Character Preprocessing}\\\hline
          \multicolumn{2}{c||}{VIPER(0.1, DCES)}   & -5 & 12 & 17 & 3 & 3 & 6\\  
          \multicolumn{2}{c||}{VIPER(0.4, DCES)}  & 5 & 20 & 23 & 8 & 10 & 13\\
          \multicolumn{2}{c||}{VIPER(0.1, ECES)} & -7 & 8 & 18 & 1 & 0 & 4\\
          \multicolumn{2}{c||}{VIPER(0.4, ECES)} & -7 & 8 & 18 & 1 & 0 & 4\\\hline\hline
          
          \parbox[t]{2mm}{\multirow{4}{*}{\rotatebox[origin=c]{90}{Grondahl}}} & Add Space & -6 & 8 & 51 & 2 & 0 & 11 \\
         & Remove Space & -6 & 8 & 66 & 34 & 0 & 20\\
         & Add `love' & 0 & 14 & 8 & 1 & 0 & 3\\
          & Remove Space, Add `love' & 0 & 14 & 64 & 38 & 0 & 23 \\\hline\hline
          \end{tabular}
    \caption{Full robustness results on OLID. Columns show accuracy drop. The approach is specified as \textit{selection - replacement} where \textit{selection = \{Greedy Select (GS), Attention Select (AS)\}} and \textit{replacement = \{\sysnamepremaxNospace, \sysnameupdtenmaxNospace\}}. 
    For our attack, the adversarial, surrogate classifier is indicated in column 1.  The first row presents baseline classification accuracies (\%) before attacks. Therefore the resulting accuracies can be calculated by subtracting the drop from the original accuracy.}
    \postcaption
    \label{tab:allResults}
\end{table*}


\presec
\section{Replication study: Reddit dataset}
\label{sec:reddit}
\postsec
We verify our initial results on a second dataset composed of 
moderated Reddit comments \cite{chandrasekharan2018internet}.
To include non-moderated comments, we collected 5.6 million comments following the same procedure as \cite{chandrasekharan2018internet}.
We then used random sampling to construct a dataset with a similar 15:1 ratio of non-moderated to moderated comments as OLID. 
The dataset has 181,519 comments split into 145,846 (4,285 moderated and 141,561 non-moderated) training comments and 35,746 (1,071 moderated and 34,675 non-moderated) test comments.
We re-build using these data and test the BERT based classifer (NULI-R) and the BLSTM Ensemble classifier (MIDAS-R).
These are tagged with a '-R' to indicate training on the Reddit dataset.
We exclude VIPER due to the previously shown weaknesses.
We also exclude the methods of \citet{grondahlLove2018} because of weak performance.

\vspace{.05in} 
\noindent
\textbf{Accuracy.}
Summarizing here, BLSTM ensemble (MIDAS-R) is most robust seeing a lower drop in accuracy, 31, than the BERT based model (NULI-R), 39, against the attack. 
Attacks using \sysnameupdtenmax, see highest drops in accuracy against MIDAS: average of 32 (range: 28 - 35), while attacks using \sysnamepremaxNospace, see highest drops against NULI (avg: 40, range: 37-42).
%
Finally, greedy select (GS) causes greater drops against NULI (avg: 40) while attention select (AS) causes greater drops against MIDAS (avg: 35). 
The results are reported in Table \ref{tab:redditResults}

\begin{table}[!t]
    \small
    \centering
    \begin{tabular}{c|c||cc}
         & & \multicolumn{2}{c}{Drop in Accuracy}\\\hline
         \multicolumn{2}{c}{No Attack Accuracy \%} & 92  & 99 \\\hline\hline 
         & Surrogate  & NULI-R  & MIDAS -R \\
         &  Classifier &  &\\\hline
         \multirow{2}{*}{GS - \sysnamepremax} & NULI-R & - & 25 \\
         & MIDAS-R & 42 & - \\\hline
          \multirow{2}{*}{GS - \sysnameupdtenmax} & NULI-R &  -  & 28\\
          & MIDAS-R & 38 & -   \\\hline
           \multirow{2}{*}{AS - \sysnamepremax} & NULI-R & - & 34 \\
         & MIDAS-R & 37 & - \\\hline
          \multirow{2}{*}{AS - \sysnameupdtenmax} & NULI-R &  -  & 35\\
          & MIDAS-R & 37 & -   \\\hline
          \end{tabular}
    \caption{Robustness results on Reddit dataset using our attack model. Cell values indicate drop in classifier accuracy. First row: classification accuracy before attack.}
    \postcaption
    \label{tab:redditResults}
\end{table}

\begin{table*}[h!]
    \small
    \centering
    \begin{tabular}{p{7.5cm}|p{7.5cm}}
         Original Comment & Adversarial Attack Comment \\
         \hline
         \hline
         you \textcolor{red}{want} me to cite statutes to prove perjury and evidence tampering are crimes are you \textcolor{red}{fucking} \textcolor{red}{retarded} & you \textcolor{red}{wants} me to cite statutes to prove perjury and evidence tampering are crimes are you \textcolor{red}{damn} \textcolor{red}{weird}  \\\hline
         you're a moron get \textcolor{red}{fucked} & you're a moron get \textcolor{red}{screwed}  \\\hline
         no you evil \textcolor{red}{cunt} & no you evil \textcolor{red}{prick} \\\hline
         she a dumb brown \textcolor{red}{bitch} & she a dumb brown \textcolor{red}{hoe}\\\hline
         you are so \textcolor{red}{fucking delusional} &  you are so \textcolor{red}{damn psychotic}
    \end{tabular}
    \precaption
    \caption{Examples to illustrate adversarial generations of Reddit comments (MIDAS-R used as surrogate classifier and \sysnameupdtenmax at the attack).}
    \postcaption
    \label{tab:obfuscExamples}
\end{table*}

%
%

\vspace{.05in} \noindent \textbf{Quality.} 
We see substitutions that subvert offense detectors such as 
\textit{trump} being replaced with \textit{trum}, which maintains the original message but now bypasses the detector\footnote{Comments on Reddit are moderated for various reasons not limited to offensive words, therefore in this case if comments against trump supporters are being moderated, it follows to change ``trump''}.
We also see errors appear, such as ``ctfu'' being substituted for ``shut''. 
Overall, results with this second Reddit dataset are consistent with OLID results underlining our conclusion that the offense classifiers are not robust against these crafty attacks.
We also see room for improvement of our adversarial attack methods especially in exploring more advanced filters for candidate substitution words.
More examples found in Table \ref{tab:obfuscExamples}.

\end{document}